# The Infinite Latent Events Model

**David Wingate, Noah D. Goodman, Daniel M. Roy and Joshua B. Tenenbaum**
Brain and Cognitive Sciences
Massachusetts Institute of Technology
Cambridge, MA 02139

## Abstract

We present the Infinite Latent Events Model, a nonparametric hierarchical Bayesian distribution over infinite dimensional Dynamic Bayesian Networks with binary state representations and noisy-OR-like transitions. The distribution can be used to learn structure in discrete timeseries data by simultaneously inferring a set of latent events, which events fired at each timestep, and how those events are causally linked. We illustrate the model on a sound factorization task, a network topology identification task, and a video game task.

## 1 INTRODUCTION

Inferring latent structure in temporal data is an important timeseries modeling problem. Consider the task of using fMRI data to recognize coherent brain activity (latent events), discover which brain activity leads to other activity (causal structure), and determine the relationship between latent brain activity and observed brain scans (observation model). As this example suggests, sequences of high-dimensional observations are often generated as the result of latent events in the external world, where events at time $t$ interact to cause events at time $t+1$. Given the observations, we wish to recover a factored representation of the latent state space and learn a structured representation of the transition probabilities governing the evolution of the state. Thus, we must *simultaneously* address the state-space learning problem and the structure learning problem of factored probabilistic models.

For various reasons, traditional state space models are inadequate for this setting. Many timeseries models descend from the Hidden Markov Model (or HMM) [13], which assumes that an unobserved Markov chain of latent states exists, and that the observation at time $t$ is conditionally independent of all other observations and states given the latent state at time $t$. A typical parametric Bayesian prior for an HMM fixes the number of latent states and places no restriction on the structure of the transition probabilities, rendering these models ineffective for data sets where the number of latent states far exceeds the amount of observed data, as would be the case in the combinatorial examples that motivate this work. The Hierarchical Dirichlet Process (HDP) HMM [1, 14] relaxes the assumption of a fixed, finite number of states, instead positing a countably infinite number of latent states and a random transition kernel where transitions to a finite number of states account for all but a tiny fraction of the probability mass. However, like parametric HMMs, the nonparametric HDP-HMM does not provide adequate inductive bias in combinatorial state spaces in the face of limited data.

This issue was one of the motivations behind the Factorial HMM [6], which assumes that the state space factors into $K$ components that evolve as independent Markov chains. The nonparametric Infinite Factorial HMM (IFHMM) [4] eliminates the need to specify $K$. While the additional structure in the Factorial HMM produces a strong inductive bias and can significantly reduce sample complexity, the independence assumption is inappropriate as we expect there to be interactions between the latent factors.

Instead, we wish to model structured causal interactions between latent events. Dynamic Bayesian Networks (DBNs) relax the independence assumption of Factorial HMMs, assuming only that the distribution of states at time $t+1$ factors when conditioned on the state at time $t$.

We present the Infinite Latent Events Model (ILEM), which can be seen as a nonparametric distribution over a restricted class of DBNs. We use the ILEM as a prior in a hierarchical Bayesian timeseries model to simultaneously address the state-space estimation problem and the structure learning problem: MAP inference in the ILEM learns a latent state sequence and its properties, including the number of relevant latent factors in the state space, the values of those factors at each timestep, and the causal structure between those factors. The ILEM combines elements of classical models and their nonparametric counterparts: we assume that observations are generated through a combination of latent events (like an IFHMM), which are coupled through a hierarchy of Dirichlet processes (like the HDP-HMM), and that there is a factored dependence structure relating successive events (like a DBN).

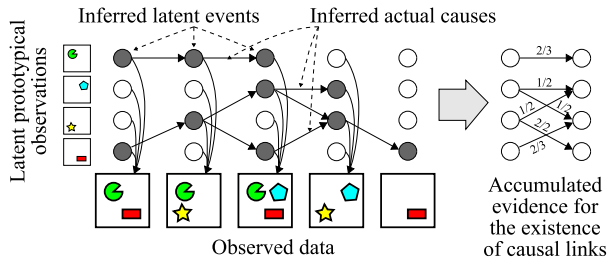

Figure 1: The ILEM can learn timeseries structure. Given observations (represented here as images) it infers latent events, actual causes, and prototypical latent observations.

The problem of structure learning in DBNs is a difficult one. Ghahramani [5], for example, assumes that the dependencies between states is known and discusses learning maximum likelihood parameter estimates of HMM, factorial and tree-structured latent state spaces using variational, sampling and EM algorithms; more recent work has focused on learning the structure itself [10], but most algorithms assume that data is fully observed [3, 2].

Our contribution is a novel perspective on what it means to learn causal structure. In contrast to other methods which explicitly attempt to describe a transition network that holds between $t$ and $t+1$ for all $t$, we leave the causal network latent, and instead reason about causality in *actual cause form:* we explicitly represent which *actual* event at time $t$ caused which *actual* event at time $t+1$, implicitly integrating out the latent causal network.

This has the effect of introducing more latent variables (there is now one for each possible cause/effect relationship at every timestep), which we couple together via a hierarchy of Dirichlet processes (DPs). The DPs allow the variables to share statistical strength, and their distributed nature softens inference and allows for emergent causal structure. Our model uses a hierarchy of DPs to capture the idea that there is always the possibility of a new dimension in the state representation that was heretofore unknown, or a new causal link in the DBN that was heretofore unused.

This addresses an important open question for many timeseries models: how to infer their dimension (e.g., the number of latent states (HMM), chains (Factorial HMM), or factors (DBN)). While the dimension of the latent space under this nonparametric prior is countably infinite, the model balances its preference to utilize as few dimensions as possible with the need to explain the data. Given data, we can determine the distribution of the number of dimensions actually used by the observed data.

## 2 THE INFINITE LATENT EVENTS MODEL

The ILEM is a nonparametric distribution over a state sequence with a latent causal structure governing state transitions. States are represented as binary vectors, with each active bit representing an *event*. Transition probabilities between states are defined by bitwise noisy-ORs, with parent events causing child events (or effects). The ILEM is a distribution $p(X, C)$ over two objects: $X$, representing events, and $C$, representing actual causes. Figure 1 illustrates this: observations are shown as images, states are represented as a binary vectors, and actual causes are represented as the arrows between events.

### 2.1 THE LATENT EVENTS

Events are represented as a matrix $X \in \{0,1\}^{T \times \infty}$, where $X_{t,i}$ indicates whether or not event $i$ was active at time $t$ (similar to the IFHMM and the Indian Buffet process (IBP) [7]). The total number of triggered events is potentially unbounded. As an example, the matrix

$$X = \begin{bmatrix} 1 & 0 & 0 & 1 \\ 1 & 0 & 1 & 0 \\ 1 & 1 & 0 & 1 \\ 0 & 1 & 1 & 0 \\ 0 & 0 & 0 & 1 \end{bmatrix} \cdots$$

represents the latent events in Fig. 1 with $T = 5$. Only four latent events are observed here.

### 2.2 THE CAUSAL STRUCTURE

To represent actual causes, we use a tensor $C \in \mathbb{N}^{T \times \infty \times \infty}$, where $C_{t,i,j}$ represents the number of times event $i$ triggered event $j$, causing event $j$ to be active at time $t$. $C$ is constrained such that $C_{t,i,j} > 0$ iff $X_{t,i} = 1$ and $X_{t+1,j} = 1$. Importantly, note that $C$ does not specify a connection topology in the usual sense—entries in $C$ should not be considered edges in a graphical model which govern state transitions. Rather, they should be viewed as *evidence* of an edge.

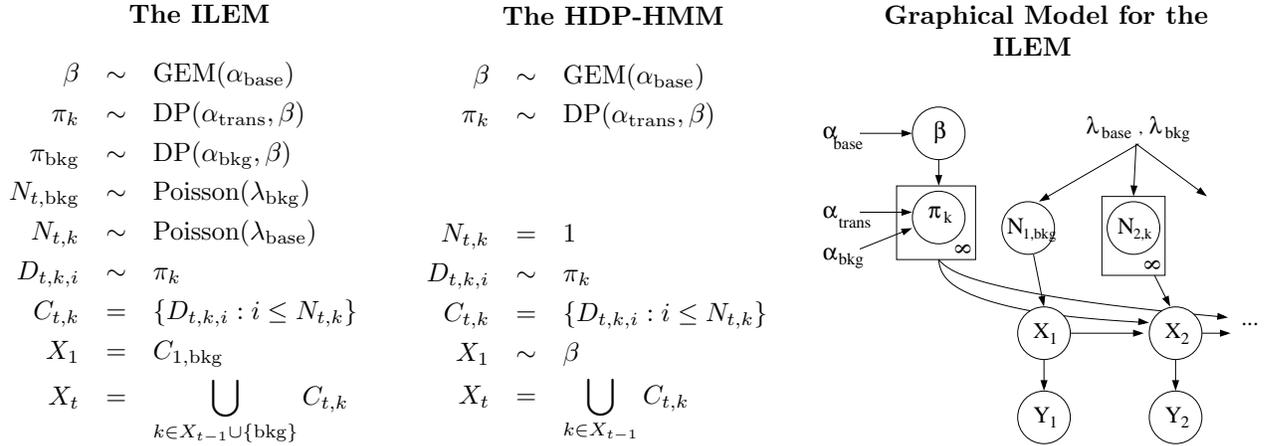

| The ILEM | The HDP-HMM | Graphical Model for the ILEM |
|---|---|---|

$$\begin{aligned}
\beta &\sim \text{GEM}(\alpha_{\text{base}}) \\
\pi_k &\sim \text{DP}(\alpha_{\text{trans}}, \beta) \\
\pi_{\text{bkg}} &\sim \text{DP}(\alpha_{\text{bkg}}, \beta) \\
N_{t,\text{bkg}} &\sim \text{Poisson}(\lambda_{\text{bkg}}) \\
N_{t,k} &\sim \text{Poisson}(\lambda_{\text{base}}) \\
D_{t,k,i} &\sim \pi_k \\
C_{t,k} &= \{D_{t,k,i} : i \leq N_{t,k}\} \\
X_1 &= C_{1,\text{bkg}} \\
X_t &= \bigcup_{k \in X_{t-1} \cup \{\text{bkg}\}} C_{t,k}
\end{aligned}$$

$$\begin{aligned}
\beta &\sim \text{GEM}(\alpha_{\text{base}}) \\
\pi_k &\sim \text{DP}(\alpha_{\text{trans}}, \beta) \\
\\
\\
N_{t,k} &= 1 \\
D_{t,k,i} &\sim \pi_k \\
C_{t,k} &= \{D_{t,k,i} : i \leq N_{t,k}\} \\
X_1 &\sim \beta \\
X_t &= \bigcup_{k \in X_{t-1}} C_{t,k}
\end{aligned}$$

Figure 2: Generative processes for the ILEM (left) and HDP-HMM (middle), and graphical model of the ILEM (right). Here $\beta$ is drawn from a GEM distribution, which is a distribution over the integers [12]; $\pi_k$ are the individual transition DPs for each possible latent event, $N_{t,k}$ are the counts of the number of child events caused by each parent event $k$ at each time $t$, and $\{D_{t,k,i}\}_{i\geq 0}$ is the set of all possible events generated by event $k$ at time $t$. The derived random variables $C_{t,k}$ and $X_t$ select the actual causes and events at every time $t$.

## 2.3 THE STOCHASTIC GENERATIVE PROCESS FOR THE ILEM

We now describe $p(X, C)$ via its generative process (see Fig. 2 for the graphical model). In the following, we assume that there is a distinguished background event, which is always active and which has its own hyperparameters, but which is otherwise identical to all other events.

Imagine that we have generated a sequence of latent states $X_1, \cdots, X_{t-1}$, recording which events caused which other events in the $C$ tensor. To generate $X_t$, we iterate over each active event $i$ at time $t-1$ and sample child events as follows:

- Sample $N_{t,i} \sim \text{Poisson}(\lambda_{\text{base}})$, which is the number of child events that $i$ actually causes at time $t$.
- Sample $N_{t,i}$ child events from a private Dirichlet process for parent event $i$.

To sample child events from event $i$'s private DP:

- Sample an event from a multinomial distribution over previously caused events. With probability $C_{i,j}/(C_i + \alpha_{\text{trans}})$, cause event $j$, which event $i$ has previously caused $C_{i,j}$ times. The total number of events previously caused by $i$ is $C_i = \sum_j C_{i,j}$.
- With probability $\alpha_{\text{trans}}/(C_i + \alpha_{\text{trans}})$, cause a new event.

To cause a new event:

- With probability $B_j/(B + \alpha_{\text{base}})$, parent $i$ causes an event that has been previously caused by other events (or the background). $B_j$ is the number of events which have ever caused $j$, so that $B = \sum_j B_j$ is the number of events caused by any latent event.

- With probability $\alpha_{\text{base}}/(B + \alpha_{\text{base}})$, parent $i$ causes a new event that has never been caused by any other event or the background.

Finally, we increment $C_{t,i,j}$ every time event $j$ triggers event $i$. We set $X_{t,i} = 1$ if $C_{t,i,j} \geq 1$ for any event $j$. (Note that $C$ can be more than 1, which could be interpreted in different ways for different causation and/or observation models; most likely, it would represent a strength of causation. We will only consider observations models that depend on $X$.)

Thus, like the HDP-HMM and other stochastic processes based on DPs, each latent event prefers to cause events which it has already caused before, but there is always some probability it will cause a new event which others have caused, or a completely new event which has never been caused before.

The parameters of the model are the concentration parameters $\alpha_{\text{base}}$, $\alpha_{\text{trans}}$ and $\alpha_{\text{bkg}}$, which govern the DPs, and the Poisson parameters $\lambda_{\text{base}}$ and $\lambda_{\text{bkg}}$.

## 2.4 PROPERTIES OF THE PARAMETERS

The ILEM can exhibit qualitatively different inductive biases depending on the hyperparameter settings. For example, the three concentration parameters $\alpha_{\text{base}}$, $\alpha_{\text{trans}}$ and $\alpha_{\text{bkg}}$ each control different DPs. For any given DP and its parameter $\alpha$, the expected number of distinct values sampled is proportional to $\alpha$.

This implies that we can control how much the model favors determinism by setting $\alpha_{\text{trans}}$ to be small. This effectively controls the number of unique successor events for any given latent factor $x_i$. We can similarly influence how many unique events are allowed to be used overall (via $\alpha_{\text{base}}$) and how many are explained by the background (via $\alpha_{\text{bkg}}$). The parameter $\lambda_{\text{base}}$ influences the expected number of child events caused by a given latent factor at any time $t$, while $\lambda_{\text{bkg}}$ similarly controls the expected number of child events caused by the background.

### 2.5 COMPUTING THE LIKELIHOOD

Given the generative process described previously, the joint probability density of a particular $X$ and $C$ is:

$$p(X,C) = \\
\left( \prod_{k=1}^{K} \text{CRP}(B_k, \alpha_{\text{base}}) \, \text{CRP}(C_{\text{bkg},k}, \alpha_{\text{bkg}}) \right) \\
\left( \prod_{i=1}^{K} \prod_{k=1}^{K} \text{CRP}(C_{i,k}, \alpha_{\text{trans}}) \right) \quad (1) \\
\left( \prod_{t=1}^{T} \text{Poisson}(N_{t,\text{bkg}}, \lambda_{\text{bkg}}) \prod_{i \in X_t} \text{Poisson}(N_{t,i}, \lambda_{\text{base}}) \right)$$

where $K$ is the total number of latent events used, active$(t)$ is the set of active events at time $t$, $N_{t,i}$ is the number of child events of parent event $i$ at time $t$, $N_{t,\text{bkg}}$ is the number of child events of the background event at time $t$, and

$$\text{CRP}(x, \alpha) = \alpha^K \frac{\Gamma(\alpha)}{\Gamma(\alpha + \sum_i x_i)} \prod_{i=1}^{K} \Gamma(x_i)$$

is the likelihood of a set of counts under a Chinese Restaurant process with concentration parameter $\alpha$.

Like other causal models, the sequence of latent events is not fully exchangeable in time; rather it is *Markov exchangeable* [1], meaning that events and their actual causes can only be exchanged if they do not change the DP counts. However, a weaker property of exchangeability is preserved: the posterior predictive transition probabilities only depend on counts of observed transitions, and not *when* those transitions occurred.

### 2.6 RELATIONSHIP TO OTHER MODELS

The ILEM is closely related to two existing dynamical system models: the HDP-HMM, and noisy-OR DBNs.

**The HDP-HMM.** The ILEM can be seen as a generalization of the HDP-HMM. For comparison purposes, Fig. 2 gives the generative model of the HDP-HMM alongside the ILEM. There are two major differences between the two models: first, the HDP-HMM does not have a background event, and second, the HDP-HMM sets $N_{t,k} = 1$. This implies that the HDP-HMM models the evolution of a single sequence of states. By sampling multiple effects for each event, the ILEM naturally extends this to modeling the evolution of a factored state sequences, with flexible causal dependencies between successive states.

**Noisy-OR DBNs.** If *any* active parent event $X_{t,i}$ triggers event $j$, then $X_{t+1,j}$ is activated. By taking the stick-breaking view on the events' private DPs, we can view the ILEM as a noisy-OR DBN with an infinite set of binary nodes at each timestep, each fully connected to the nodes at the next timestep. Using the CRP representation of the DP implicitly integrates out all of these connections, leaving just a few hyperparameters. This relationship to noisy-OR DBN's is formalized in Appendix A; similar connections between DPs, IBPs and noisy-OR models are described by Wood et al. [16].

### 2.7 OBSERVATION MODELS

To complete the generative model, we must specify an observation model $p(Y|X,C)$. We now describe one possible choice: the linear-Gaussian observation model used by Griffiths & Ghahramani [7].

In this model, an observation $y_t$ is generated from a Gaussian with mean $\mathbf{A}x_t$ and covariance matrix $\Sigma_Y = \sigma_Y^2 \mathbf{I}$. Assuming $D$-dimensional observations, the matrix $\mathbf{A}$ is $D \times K$ (where $K$ is the maximum number of events which are ever used). Each column $A_i$ can be thought of as a prototypical observation associated with latent event $x_i$; the observation $y_t$ is a linear combination of these prototypical observations plus noise. In an image modeling context, for example, each column $A_i$ would correspond to a prototypical sprite associated with latent event $x_i$.

The prototypical representations are unknown, and so the model places a matrix Gaussian prior (with variance $\sigma_A^2$) on $\mathbf{A}$, and then integrates out $\mathbf{A}$. This yields the following marginal density:

$$p(\mathbf{Y}|\mathbf{X}) = \frac{1}{Z} \exp\left\{ -\frac{1}{2\sigma_Y^2} \text{tr}(\mathbf{Y}^T(\mathbf{I} - \mathbf{X}\mathbf{M}^{-1}\mathbf{X}^T)\mathbf{Y}) \right\}$$

where $Z = (2\pi)^{\frac{ND}{2}} \sigma_Y^{(N-K)D} \sigma_A^{KD} |\mathbf{M}|^{\frac{D}{2}}$ is the Gaussian normalization constant and $\mathbf{M} = \mathbf{X}^T\mathbf{X} + \frac{\sigma_Y^2}{\sigma_A^2}\mathbf{I}$. Under this observation model (see [7]), the means of the Gaussians associated with each latent event can be recovered in closed form as:

$$\mathbb{E}[\mathbf{A}|\mathbf{Y},\mathbf{X}] = (\mathbf{X}^T\mathbf{X} + \frac{\sigma_Y^2}{\sigma_A^2}\mathbf{I})^{-1}\mathbf{X}^T\mathbf{Y}$$

Later, we use this equation to recover the prototypical latent observations associated with each event.

## 2.8 INFERENCE

We now describe our MCMC inference algorithm for the ILEM using the linear-Gaussian observation model (see Sec. 2.7).

To begin, we make a simplifying approximation. It is the case that the linear-Gaussian observation is conditionally independent of the actual cause matrix $C$ given the the binary event matrix $X$. Specifically, how many times event $i$ triggers event $j$ has no effect on the likelihood—only whether the event was triggered or not affects the observation. To make inference efficient, we constrain our sampler to explore only the space where $C$ is binary-valued, sampling, in effect, from a conditional distribution of the model. We still score with Eq. 1, which is now only proportional to the joint density; this presents no difficulty for the MCMC sampler.

In order to explore the latent space, we use a combination of single-site Gibbs sampling steps and Metropolis-Hastings (MH) moves. In particular, we sample each entry in the event matrix and actual cause matrix using Gibbs updates, because the conditional distributions have trivial closed forms given our binary assumption above. Changes to $X$ and $C$ only change a few sufficient statistics within each DP and Poisson, so changes to the likelihood can be computed incrementally. Often, such incremental changes can be propagated to the observation model; in the case of the linear-Gaussian model, efficient rank-one updates can be applied to the necessary matrix inverses and determinants.

We found that certain MH moves improved mixing considerably. One particularly useful MH move randomly rewrites a candidate event $X_{t,i}$ as a different event $X_{t,k}$, rewiring causes and effects to the event's new label. For this move, we consider all possible rewrites for all active latent events at a particular timestep. Another useful MH move swaps parent events, turning off $C_{t,i,k}$ and turning on $C_{t,j,k}$ simultaneously. Finally, we annealed the sampler during burn-in using a geometric temperature schedule. For MAP estimates, we continued to anneal the temperature towards zero.

## 3 EXPERIMENTS

We evaluate the ILEM on three different data sets designed to illustrate different properties of the model: a causal soundscape factorization task; a simple video game task, and a network topology discovery task. In each, our goal is to identify the maximum aposteriori parameter estimates of $p(X, C|Y) \propto p(Y|X, C)\, p(X, C)$.

### 3.1 CAUSAL SOUNDSCAPE FACTORIZATION

The linear-Gaussian observation is appropriate in situations where latent prototypical observations combine linearly to produce a composite observation, which is true of sound data. This is illustrated graphically in the top of Figure 3. For this experiment, we constructed a simple "jungle soundscape," in which different jungle sounds caused other jungle sounds: jaguars howling caused birds to twitter, elephants trumpeting caused frogs to croak, crickets tended to stay on or off for long periods of time, etc. 52 observations were generated, each of which was an 8K sound sample. This problem is closely related to blind source separation tasks, except that we are additionally interested in the latent causal structure.

Figure 3 shows the results. The bottom left shows the ground truth for which events were active at which times, and the bottom right shows the learned events and causal structure. In this case, the ILEM correctly discovers the latent dimensionality of the data, the causal network, and an almost perfect set of latent events.

We chose this problem to illustrate a simple example where the HMM would fail to find the correct structure because it was unable to represent the factored nature of the state space and learn from only a limited amount of data. A factorial HMM with independent binary-valued Markov chains would also be unable to represent the structure of the data because different events interact.

### 3.2 SPACE INVADERS

Next, we experimented on a stylized version of space invaders. A single alien travels back and forth while a single turret moves back and forth, shooting bullets at the alien. If the bullet hits the alien, the alien explodes and resets to an initial position. Figure 4 on the left shows typical images from the data set.

This problem was chosen to illustrate the need for both factored state representations and flexible causal structure. In this case, we expect the HMM to fail because of the combinatorics of the latent space, and we expect a factorial HMM to fail because the different latent factors interact with one another.

There are approximately 25 latent events in the data set: there are six positions for the alien, three for the turret, nine different bullet positions, three different explosion positions, and four action indicators (represented as bars on the bottom of the image). Observations are 15x15 binary images. The ILEM was trained on 400 observations. A good initialization greatly accelerated learning, so we initialized with a non-negative matrix factorization algorithm [11], which initialized both the $X$ matrix and the $A$ matrix with 30 different events. Of the 30 events, 13 of them did not correspond neatly to a single object – they lumped multiple objects or parts of objects together. It was then up to the ILEM to clean this up, by determining exactly how many objects there were, and infer the causal structure relating them.

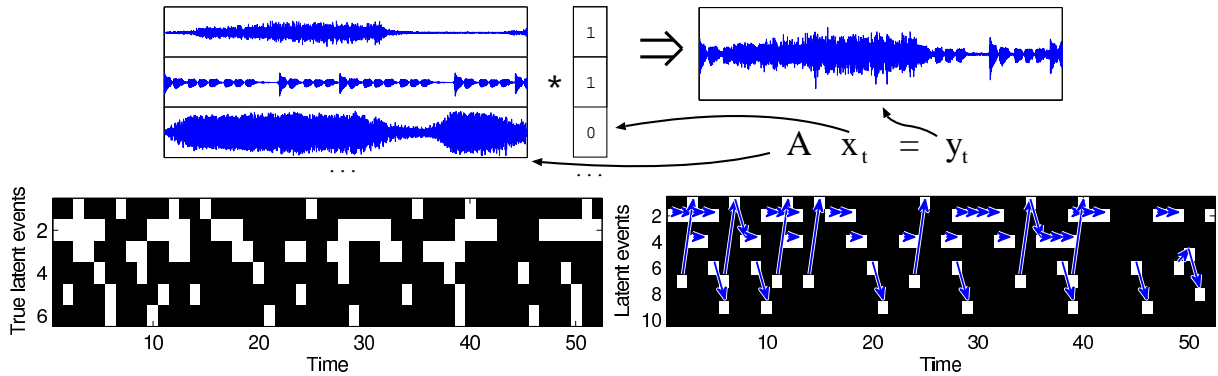

Figure 3: On the top: setup for the causal soundscape factorization task. Sound events are linearly combined to form a composite observation. On the bottom left: the ground truth showing which events were active at which times. Each row represents a single sound, such as a bird, elephant, tiger, etc. On the bottom right: the inferred latent events and actual causes. Cricket (row 2) and frog (row 4) sounds evolve independently, lions roaring (row 7) cause birds to screech (row 1), elephants trumpeting (row 6) cause monkeys to howl (row 9), etc.

We evaluated the results in two ways. First, we examined the learned prototypical observations associated with the latent events. Several of these prototypical observations are shown in the upper-right of Fig. 4. They correspond to objects in the world: an event for the explosion in each position (top row), different events for the different positions of the bullets (middle row), and different events for the alien (bottom row).

We also examined the inferred causal network. On the bottom of Fig. 4 are two chains of causal structure. The top shows the alien looping from the upper-left to the upper right and back again. The model successfully distinguishes the fact that the alien stays in the upper-right corner for two successive timesteps (although it does not make the same prediction for the upper-left corner). Just below is shown the progression of events as a bullet is fired from the middle position.

### 3.3 NETWORK TOPOLOGY DISCOVERY

For these experiments, the ILEM was tasked with discovering a causal network topology based loosely on the SysAdmin problem [8]. In the SysAdmin problem, a sysadmin must keep a network of computers working. Machines crash periodically; when they do, they cause a cascade of failures, which is governed by a network topology. Our version of the problem is to discover the network topology, given only the crash data.

We chose this problem to clearly illustrate how we can infer latent nodes in a DBN, while also inferring the causal structure of the network. We again expect the IFHMM and HMM to fail for reasons similar to those discussed previously.

We represent the data as a binary matrix with $T$ rows, and $K$ columns, where $T$ is the number of timepoints, and $K$ is the number of machines, and $X_{t,k} = 1$ iff machine $k$ has failed at time $t$.

**First variant: known machines.** For this variant, the number of latent events is known – we have one for each machine. We therefore fix that parameter of the ILEM, and only attempt to infer the network topology. (A nonparametric approach is still useful because we do not know the number or strength of the edges.)

We tested the ILEM on four topologies: the three shown in Figure 5, and a special fourth topology (explained below). For the three small topologies, we used 400 data points. The ILEM discovers almost perfect topologies: for the four-node ring, the ILEM correctly identifies 12/12 of the existing and missing links; for the five-node star, it identifies 19/20 existing and missing links; for the seven-node tree, the ILEM identifies 40/42 existing and missing links. Note that the matrices of one-step co-occurrence statistics is often largely uninformative, as shown in the bottom of Figure 5. In order to discover a reasonable causal network, we need the ILEM's bias towards a sparse connectivity.

For the fourth topology, we generated a randomly connected network among 100 machines (with about 99.5% sparsity), and generated 1,000 data points. The best network derived from the posterior ILEM correctly inferred 9,982/10,000 existing and missing links; in contrast, the best network inferred from co-occurrence statistics missed twice that many.

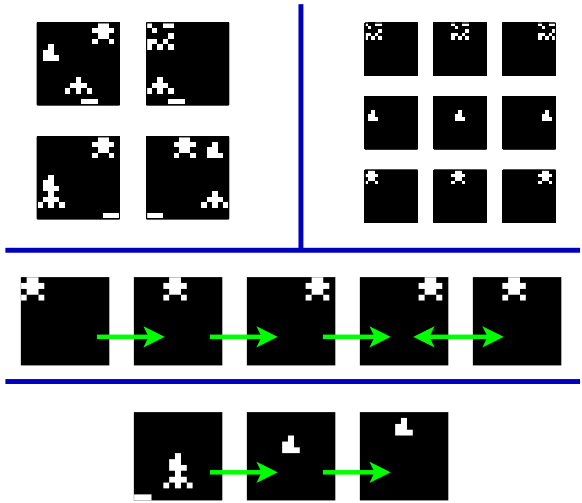

Figure 4: Results on space invaders. Top left: typical observations. Top right: several of the prototypical observations associated with each event: the top row shows explosions, the middle row shows some bullet events, and the bottom row shows the alien. Bottom figure: two chains of causal events inferred by the model. The alien moves back and forth, and a bullet moves upward after being fired from the ship.

**Second variant: unknown machines.** For this variant, we hid one of the machines in each small network (modeling a rogue machine that the system administrator is unaware of; shown as a shaded node in Fig. 5), and attempted to infer a complete network. Now, both the number of machines and the topology are unknown. The results were encouraging: for all three networks, the ILEM was able to correctly infer that a single rogue machine existed, and was able to infer almost perfect network topologies; there were more extraneous entries in the $C$ tensor, representing more uncertainty about the exact underlying network, but the correct links almost always had the strongest evidence for their existence.

## 4  CONCLUSIONS

We have presented the Infinite Latent Events Model, a nonparametric Bayesian model capable of capturing factored causal structure in timeseries data with rich observations. The centerpiece of our model is an actual-cause approach which leaves the DBN structure latent. Using a Gibbs sampler to perform inference of a hierarchy of Dirichlet processes, we have shown that the model can infer an appropriate set of latent events

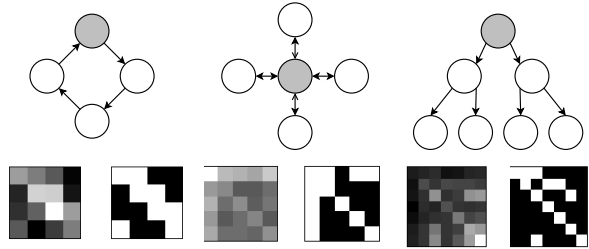

Figure 5: Setup and results for the SysAdmin problem. Top: network topologies tested. Bottom: co-occurrence statistics (left) and the ILEM's learnt networks (right).

and the causal structure connecting them.

Efficient inference in the ILEM on large data sets remains an important open research problem; we believe that some combination of slice sampling and dynamic programming—like that employed for the IFHMMs (see [4])—and/or particle filtering (e.g., Beal's infinite state particle filter [1]) are promising options. From a modeling perspective, inhibitory (and more general) causal link models, hierarchical structure, and semi-Markov extensions appear to be straightforward ways to extend the applicability of the ILEM to a wider range of new data. For many timeseries data, especially those acquired by autonomous agents, we believe that natural models will require a notion of objects; a flexible prior with such an inductive bias is an important next step.

## A  RELATION TO NOISY-OR DBNs

The active events $X_t$ at time $t$ are given by the union over Poisson independent draws from $\pi_k$ for each active event $k \in X_{t-1}$. Let $Q$ denote the conditional distribution of $X_t$ given $X_{t-1}$ (treating the $\{\pi_k\}$ distributions as nonrandom). We show that $Q$ is equivalent to the marginal distribution of a random, infinite-dimensional noisy-OR DBN. For each event $i \geq 0$, let $p_i = (p_{i \to 1}, p_{i \to 2}, \dots)$, where $p_{i \to j}$ is the probability that event $i$ causes event $j$ at the next timestep, given that event $i$ is currently active.

Let $X_t \subseteq \mathbb{N}$ be the set of events active at time $t$. For each event $j \in \mathbb{N}$ at time $t$, let the probability that event $j$ is active (i.e. $j \in X_t$) be

$$1 - \prod_{i \in X_{t-1} \cup \{0\}} (1 - p_{i \to j}),$$

where we have assumed that a distinguished background event $i = 0$ is always active. The above alternative conditional distribution of $X_t$ given $X_{t-1}$ is a noisy-OR DBN parameterized by $\{p_i\}_{i \geq 0}$.

**Theorem A.1.** *There is a distribution on probabilities $\{p_i\}_{i \geq 0}$ such that the conditional distribution of $X_t$ given $X_{t-1}$ is $Q$.*

*Proof.* We will show that a transformed beta process prior [9] on $p_i$ achieves the desired result. Let $\{q_x\}_{x \in \mathbb{R}} \sim \text{BP}(c, \lambda B_0)$ be a draw from a beta process with concentration parameter $c = 1$, mass parameter $\lambda = \lambda_{\text{base}}$ and continuous base distribution $B_0(x)$ on $\mathbb{R}$. Then $q_x \in [0, 1]$ for all $x \in \mathbb{R}$ with probability one.

The set of random probabilities $\{q_x\}_{x \in \mathbb{R}}$ have the following relevant properties [15]: Let $F = \{x \in \mathbb{R} : q_x > 0\}$ be the set of $x \in \mathbb{R}$ assigned positive probability. Then $F$ is a random, countably infinite subset of $\mathbb{R}$. Furthermore, the elements of $F$ are themselves i.i.d. draws from $B_0$ and independent of the probability values $q_x$. Finally, let $G \subset \mathbb{R}$ be the random set of successes among independent Bernoulli($q_x$) trials for each $x \in F$. Then the mass parameter $\lambda$ is the mean number of successes $|G|$. In fact, the number of successes has a Poisson($\lambda$) distribution.

Fix an event $k \geq 0$ and let $\pi_k$ be the corresponding measure on events and let $B_0$ be the uniform distribution on $[0, 1]$. There exists a measurable mapping $\rho_k : \mathbb{R} \to \mathbb{N}$ such that $B_0 \circ \rho_k^{-1} = \pi_k$. Because each $x \in G$ is i.i.d. $B_0$, $(\rho(x))_{x \in G}$ are Poisson($\lambda$)-many independent draws from $\pi_k$. This describes $Q$ and, furthermore, this process is equivalent to sampling from the noisy-OR network with probabilities $p_{k \to j} = 1 - \prod_{x \in \rho^{-1}(j)} (1 - q_x)$. □

## Acknowledgments

We thank the anonymous reviewers for their helpful comments. The authors also gratefully acknowledge the financial support of NTT Communication Sciences Lab, AFOSR (grant #FA9550-07-1-0075), ONR (grant #N00014-07-1-0937), NSF (GRFP for D. Roy), ARO (grant #W911NF-08-1-0242), and the James S. McDonnell Foundation Causal Learning Collaborative.